# How much should you ask?
# On the question structure in QA systems.

Dominika Basaj[*1], Barbara Rychalska[*2], Przemysław Biecek[3], and Anna Wróblewska[4]

d.basaj, b.rychalska, p.biecek, a.wroblewska@mini.pw.edu.pl
[1,2,3,4]Warsaw University of Technology, Warsaw, Poland
[1]Tooploox, Wrocław, Poland

## Abstract

Datasets that boosted state-of-the-art solutions for Question Answering (QA) systems prove that it is possible to ask questions in natural language manner. However, users are still used to query-like systems where they type in keywords to search for answer. In this study we validate which parts of questions are essential for obtaining valid answer. In order to conclude that, we take advantage of LIME - a framework that explains prediction by local approximation. We find that grammar and natural language is disregarded by QA. State-of-the-art model can answer properly even if 'asked' only with a few words with high coefficients calculated with LIME. According to our knowledge, it is the first time that QA model is being explained by LIME.

## 1 Introduction

Release of SQuAD (Rajpurkar et al., 2016) dataset boosted development of state-of-the-art solutions in Question Answering (QA) systems. Questions asked in natural way give opportunity for human-computer interaction. However, in real life scenario, users are used to 'querying' rather than 'asking'. This assumption inspired us to investigate whether QA systems trained on SQuAD dataset could be used by people who prefer to write faster and more intuitive queries. Our experiments indicate that indeed, QA system returns true answer once we type in just selected keywords without keeping the sentence structure. We conclude that QA systems have a very limited understanding of natural language. They rather learn to distinguish specific words. This indifference to semantics-altering edits is called overstability (Jia and Liang, 2017). Research on this issue was also recently conducted by Mudrakarta et al. (2018) who compute importance of words by application of Integrated Gradients.

Inspired by LIME (Ribeiro et al., 2016) we perturb questions and score newly created examples with context held constant. We prove that we can remove up to over 90% of words in question and still get the right answer.

The contribution of our study is the following:

1. We use LIME in QA model for determining which parts of question are substantial for obtaining right answer. We obtain valid results although QA systems are not natural candidates for explanation with LIME, since they do not solve a regular classification problem.

2. We show that QA models disregard grammar and syntax and thus can give the right answer once queried with most important keywords, which are the words with high coefficients returned by LIME. Our findings can serve as a starting point for development of QA models that are immune to adversarial examples and as a result - generalize better.

We use QA system developed by Chen et al. (2017). We pick this model for its good performance combined with simplicity and popularity of the algorithm, which in its basic form builds the core of many other QA models.

## 2 Experiments

In order to query QA system with most important words indicated by coefficients estimated with LIME, we design a two-step algorithm. **First**, we adjust the logic of LIME to our problem, which is perturbing questions while holding the context unchanged. We treat each word in context as a separate class and words in question serve as features. This way, we run LIME in a multiclass setting, with a varied number of c̀lasses"for each run.

---
*Both authors contributed equally.

| Question | Answer |
|---|---|
| What type of rock is found at the Grand Canyon? | sedimentary |
| type of rock Grand Canyon | sedimentary |
| type | sedimentary |

Table 1: Questions and answers after removing important words.

We inspect coefficients estimated for ground truth class (first word in answer). **Second**, once we estimate the influence of each feature, we iteratively remove one word starting with lowest coefficients. After each removal we ask reduced question until we are left with only one word. We call the shortest form of question that still gives the right answer a *root question*. We treat as a right answer a returned span of tokens in which we can locate at least one word from ground truth i.e: *question*: To promote accessibility of the works, what did Luther remove? *ground truth*: impediments and difficulties *QA answer*: impediments and difficulties so that other people may read it without hindrance.

We inspect 800 examples, analyzing questions for which the QA system predicted right answers.

**Results.** Table 1 presents example of algorithm performance. In this particular case we observe that by leaving only one word *type* we still get the right answer. As shown in figure 1 this word has the highest LIME coefficient. This is quite surprising as one-word question does not convey sufficient information about what we want to ask. Figure 2 shows distribution of percentages of removed words from question that do not disturb the answer. It is left-skewed indicating that a large proportion of question can be removed. We observe that root questions consist mostly of *wh*-words and nouns, as displayed in table 2.

This behavior can be partly traced down to the characteristics of SQuAD dataset. Due to the shortness and focus on just single topic in contexts, the network needs just a single keyword to infer the likely remainder of the question. For example, if we query "type" in a text about rocks in Grand Canyon, it is almost guaranteed that context mentions just single "type" which refers to the rock itself.

Moreover, we observe that there are questions which start off with wrong answer, but when we remove one or more words they start to consistently give valid answers. Based on this, we can hypothesize that although performance of QA systems does not depend on grammar, there are still

| Word/PoS/Phrase | % occurences |
|---|---|
| *wh*-word + 0 or more words (any) | 51 % |
| 1 word (any) | 32 % |
| 1 noun | 18 % |
| *who* | 16 % |
| *wh*-word + 1 word (any) | 13 % |
| *what* | 12 % |
| 7 and more words | 10 % |

Table 2: Most common words and phrases found in root questions.

some underlying dependencies between words.

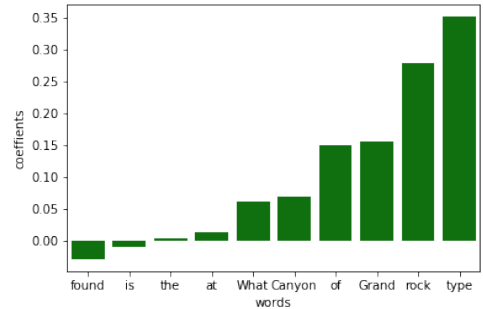

Figure 1: LIME coefficients estimated per word.

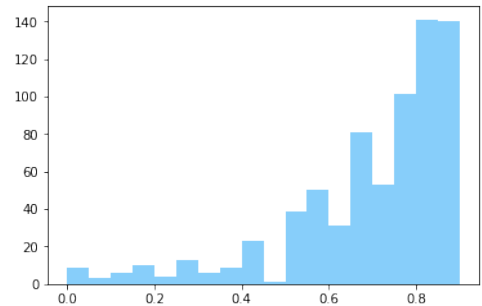

Figure 2: Distribution of percentages of removed question words that still give a valid answer.

## 3 Summary

In this study we show that QA models do not need grammar to answer questions correctly once they are left with keywords. It indicates that actually model does not really encode what we want to ask, but rather recognizes specific words and associates them with the answer. It means that words that we as humans perceive as important part of questions are disregarded in reality. This might indicate a problem with the underlying dataset, as root questions contain too little information to be considered valid in a real world setting. Our study sets a direction for decreasing their overstability by highlighting drawbacks of QA models.